\pdfoutput=1

\documentclass[11pt]{article}

\usepackage[final]{acl}

\usepackage{times}
\usepackage{latexsym}

\usepackage[T1]{fontenc}

\usepackage[utf8]{inputenc}

\usepackage{microtype}

\usepackage{inconsolata}
\usepackage{graphicx}
\usepackage{multirow, booktabs}
\usepackage{amsmath}
\usepackage{stmaryrd}
\usepackage{amssymb}
\usepackage{bbm}
\usepackage{bm}
\usepackage{enumitem}
\usepackage{subcaption}
\usepackage{xcolor}

\newcommand*\concat{\mathbin{\|}}
\title{Identifying while Learning for Document Event Causality Identification}

\author{Cheng Liu\textsuperscript{\textnormal{1}},
	Wei Xiang\textsuperscript{\textnormal{2}},
	Bang Wang\textsuperscript{\textnormal{1,}}\Thanks{\ Corresponding author},
	\\
	\textsuperscript{1}School of Electronic Information and Communications, \\ Huazhong University of Science and Technology, Wuhan, China \\
	\textsuperscript{2}School of Software Engineering, \\ Huazhong University of Science and Technology, Wuhan, China \\
	\texttt{\{liu\_cheng, xiangwei, wangbang\}@hust.edu.cn}}

\begin{document}
\maketitle
\begin{abstract}
Event Causality Identification (ECI) aims to detect whether there exists a causal relation between two events in a document. Existing studies adopt a kind of \textit{identifying after learning} paradigm, where events' representations are first learned and then used for the identification. Furthermore, they mainly focus on the causality existence, but ignore causal direction. In this paper, we take care of the causal direction and propose a new \textit{identifying while learning} mode for the ECI task. We argue that a few causal relations can be easily identified with high confidence, and the directionality and structure of these identified causalities can be utilized to update events' representations for boosting next round of causality identification. To this end, this paper designs an \textit{iterative learning and identifying framework}: In each iteration, we construct an event causality graph, on which events' causal structure representations are updated for boosting causal identification. Experiments on two public datasets show that our approach outperforms the state-of-the-art algorithms in both evaluations for causality existence identification and direction identification.\footnote{\ The source code is available at \url{https://github.com/LchengC/iLIF}}
\end{abstract}

\section{Introduction}
Event Causality Identification (ECI) is the task of identifying whether there exists a causal relation between two events. ECI can facilitate a wide range of practical applications, including knowledge graph construction~\citep{chen2019uhop,al2020end}, question answering~\citep{oh2017multi}, and information extraction~\citep{xiang2023survey}. The ECI task can be divided into the sentence-level ECI (two events are in the same sentence) and document-level ECI (two events may be in different sentences).


\par
In this paper, we focus on the document-level ECI task, which faces greater challenges due to the requirement of comprehending long texts for cross-sentence reasoning. The traditional feature-based methods~\citep{gao2019modeling} utilize Integer Linear Programming (ILP) to model the document causal structure. In order to better capture the interactions among events, recent methods~\citep{phu2021graph,chen2022ergo,chen2023cheer,fan2022towards} usually construct document-level undirected graphs to facilitate cross-sentence causal reasoning. Other methods~\citep{yuan2023discriminative} use sparse attention to address the issue of long-distance dependencies and distinguish between intra- and inter-sentential reasoning.

\begin{figure}[t]
	\centering
	\includegraphics[width=0.95\columnwidth]{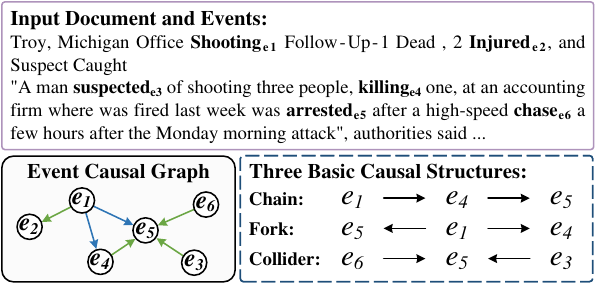}
	\caption{An example of the event causality graph and event structures in the EventStoryLine corpus. }
	\label{Fig:Introdcution}
\end{figure}

\par
Modeling the interactions among events has been proven effective for the document-level ECI task, however, almost all existing methods focus on only identifying the existence of causal relation between the event $e_i$ and $e_j$, yet without considering the causality direction being from $e_i$ to $e_j$ (or from $e_j$ to $e_i$). In this paper, $e_i \rightarrow e_j$ indicates that ``event $e_i$ causes $e_j$''. This may lead to the learning of events' representations towards capturing events' correlations, but correlations may not be directly mapped into causalities~\citep{pearl2018book}. Furthermore, undirected connections may also lead to incorrect causality identifications, as some properties of causal structures cannot be respected without directionality.

\par
There are three basic causal structures, namely, the chain, fork, and collider~\citep{he2021daring}. Causality identification without directionality cannot well exploit causal structures. As shown in Figure~\ref{Fig:Introdcution}, ``\textit{Shooting\textsubscript{e1}}''$\xrightarrow{}$``\textit{killing\textsubscript{e4}}''$\xrightarrow{}$``\textit{arrested\textsubscript{e5}}'' is a chain causal structure. For a model considering causality direction, if the two directional causal relations, i.e., $e_1 \rightarrow e_4$ and $e_4 \rightarrow e_5$, can be first identified with high confidence, then this can help to identify the causal relation between $e_1$ and $e_5$ due to the causal transmission in the chain structure. We argue that events' causal relation should be with directionality, and considering causal directions could further boost event causality identification.

\par
Besides ignoring directionality, existing solutions for the ECI task adopt a kind of \textit{identifying after learning} paradigm. That is, learning events' representations first via some advanced neural networks, and then identifying causal relations for all event pairs at only one pass. However, it could happen that some causal relations can be easily identified with high confidence. As reported by~\citet{yuan2023discriminative}, identifying intra-sentence events' causality (two events in a same sentence) is often easier and with better accuracy than identifying inter-sentence events' causality (two events in different sentences). This motivates us to propose a new \textit{identifying while learning} mode for the ECI task. That is, identifying some events' causal relations with high confidence, and then utilizing the directionality and structure of such identified causalities to update events' representations for boosting next round of causality identification.

\par
Motivated from the aforementioned considerations, this paper proposes an \textit{iterative Learning and Identifying Framework} (iLIF) for the document-level event causality identification. For an event $e_i$, we not only encode its contextual text representation $\mathbf{h}_i$, but also update its causal structure representation $\mathbf{z}_i$ in each iteration. Causality identification is modeled as a classification issue based on the representation $\mathbf{h}_i$ and $\mathbf{z}_i$ of an event pair. Initially, we employ a pretrained language model to encode $\mathbf{h}_i$. In each iteration, we first construct a directed \textit{event causality graph} (ECG) based on the identified causalities, and propose a causal graph encoder to next update $\mathbf{z}_i$ on the ECG. After the termination, we output the directed ECG as the final causality identification results. In order to differentiate the importance of iterations, we design a novel iteration discounted loss function to mitigate the error propagation issue.

\par
We conduct experiments on two public datasets: The EventStoryLine(v0.9) dataset~\citep{caselli2017event} and MAVEN-ERE dataset~\citep{wang2022maven} and consider both direction and existence settings for causal relations. We preprocess the EventStoryLine dataset~\footnote{See Appendix~\ref{app:ESCPreprocessing} for more details. } to ensure that each ground truth ECG is a directed acyclic graph~\citep{gopnik2007causal}. Experiment results validate that our iLIF outperforms the state-of-the-art competitors for the document-level ECI task in evaluations for both causality existence identification and direction identification.

\section{Related work}
\paragraph{Sentence-level ECI} Early methods in feature engineering construct classifiers by searching for effective features, such as connective word categories~\citep{zhao2016event}, syntactic features~\citep{pitler2009automatic}, and contextual semantic features~\citep{do2011minimally}. Some studies have employed external knowledge bases or linguistic tools. \citet{cao2021knowledge} induce descriptive knowledge and relation path knowledge from the ConceptNet for reasoning. \citet{liu2021knowledge} enhance model recognition capability by mining context-specific patterns from the ConceptNet. \citet{zuo2020knowdis} design data augmentation methods based on lexical knowledge bases like the WordNet~\citep{fellbaum1998wordnet} and VerbNet~\citep{schuler2005verbnet} to generate more training data. \citet{zuo2021improving} introduce specific causal patterns and transfer them to the target model using a contrastive transfer learning framework. \citet{hu2023semantic} utilize an AMR parser~\citep{banarescu2013abstract} to transform text into semantic graphs, enabling explicit semantic structure modeling and implicit association mining. \citet{shen2022event} leverage the prompt learning paradigm to jointly construct derived templates in order to utilize latent causal knowledge.

\begin{figure*}[h]
	\centering
	\includegraphics[width=0.95\textwidth,height=0.48\textwidth]{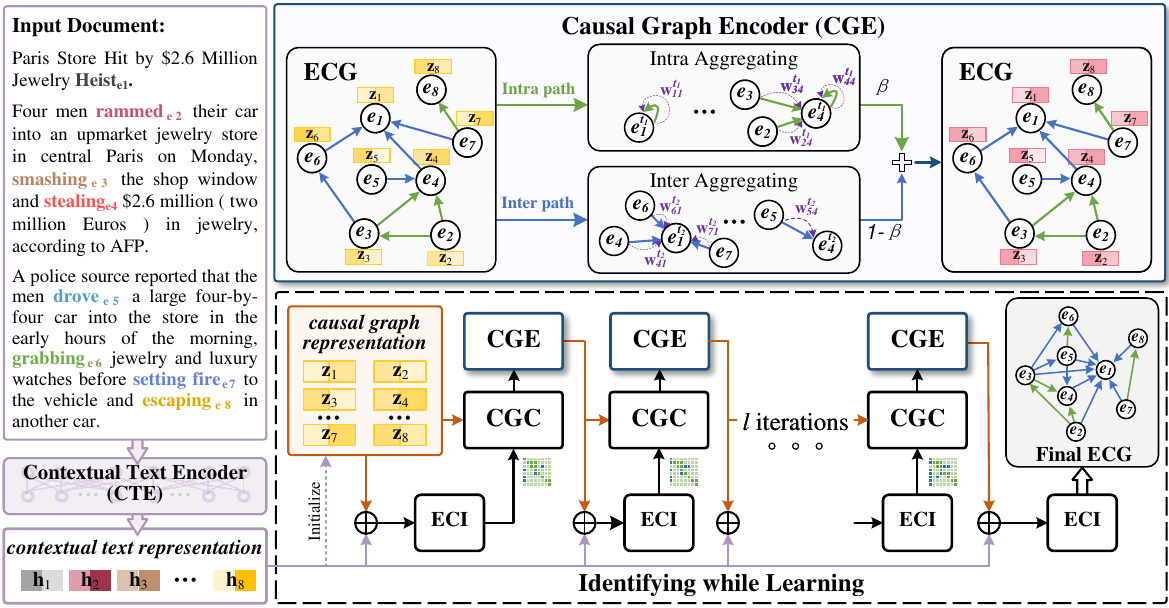}
	\caption{Illustration of the iterative learning and identifying framework (iLIF). Causality identification is based on events' contextual text and causal graph representations. The event causality graph (ECG) is iteratively constructed to update events' causal graph representations. The final ECG contains all identified causal relations as the output. }
	\label{fig:model}
\end{figure*}

\paragraph{Document-level ECI} Compared to the sentence-level ECI, document-level ECI holds greater potential for applications but also faces greater challenges, such as weak long-term dependencies and a lack of clear causal indicators. To address these challenges, researchers attempt to construct the topological structure of events for global reasoning. \citet{gao2019modeling} propose an approach based on Integer Linear Programming (ILP) to model the global causal structure. \citet{phu2021graph} construct a document-level event graph and utilize GNN to learn structural features. \citet{chen2022ergo} build a relational graph and modeled the interactions between event pairs. Furthermore, \citet{chen2023cheer} introduce prior knowledge such as central events and coreference to construct an event interaction graph for achieving global reasoning. \citet{yuan2023discriminative} propose a method that utilizes sparse attention to learn high-quality representations and distinguish between intra- and inter-sentential reasoning.

\par
To the best of our knowledge, existing methods ignore casuality direction and adopt the identifying after learning paradigm for the ECI task. In this paper, we take care of causality direction and propose an identifying while learning mode.

\section{Methodology}\label{Sec:Methodology}
We propose an \textit{iterative Learning and Identifying Framework} (iLIF) for document-level event causality identification. The basic idea is to iteratively update events' representations for causality identification by exploiting causal structures on the most recent event causality graph. As shown in Figure~\ref{fig:model}, the iLIF includes four main module: (1) \textit{Contextual Text Encoder} (CTE); (2) \textit{Causal Graph Encoder} (CGE); (3) \textit{Event Causality Identification} (ECI) (4) \textit{Causality Graph Construction} (CGC).

\subsection{Contextual Text Encoder}\label{Sec:Contextual Text Encoder}
The CTE module is to encode contextual text information for each event mention. Given a document $\mathcal{D}$ with $n$ sentences and the $j$-th sentence ${S_j}$ contains $m$ words. We use a Pretrained Language Model (PLM), say the BERT-base~\citep{devlin2018bert}, to sequentially encode the $n$ sentences and output the encodings for all the words in the document $\mathcal{D}$. For the $i$-th event $e_i$, its \textit{contextual text representation} $\mathbf{h}_i$ is an average of individual token representations of the event trigger words.

\subsection{Causal Graph Encoder}\label{Sec:CausalGraphEncoder}
Based on the contextual text representation $\mathbf{h}_i$, we can apply a simple neural classifier, such as a multi-layer perceptron, to identify the existence of a causal relation between two events. Although this naive approach is often not with excellent performance, some identified causal relations may be with high confidence. So we can choose some identified causal relations with high confidence to construct a \textit{event causality graph} (ECG), denoted by $\mathcal{G}$. We will discuss how to construct and update $\mathcal{G}$ in the next subsection.

\par
Besides describing the causality of two events, the ECG $\mathcal{G}$ contains more information about causal relations for all events in a document. Furthermore, some causal structures in $\mathcal{G}$ can be used to boost inferring new causal relations (such as the chain/fork causal structure) or to help correcting unreasonble causal relations (such as the collider causal structure). To encode causal direction and structure information from $\mathcal{G}$, we propose to learn a \textit{causal graph representation} for each event node, denoted by $\mathbf{z}_i$ (with potentially different cardinality F).

\par
We construct $\mathcal{G}$ as a heterogeneous directed graph containing one type of event node and two types of directed edges: Intra-sentence and Inter-sentence. We first utilize the self-attention~\citep{vaswani2017attention} on the event nodes using a shared attentional mechanism $\mathsf{attention}:\mathbb{R}^{F} \times \mathbb{R}^{F}\rightarrow\mathbb{R}$
to compute the type-specific importance coefficient~\citep{velickovic2017graph}. For two events $e_i$ and $e_j$ with an edge $r_{ij}^t$ of type $t$, the type-specific importance coefficient $c_{ij}^t$ of $r_{ij}^t$ is computed as follows:
\begin{align}\label{Eq:AttentionCoefficient}
	c_{ij}^t=\mathsf{attention}(\mathbf{W}\mathbf{z}_i, \mathbf{W}\mathbf{z}_j \; \text{;} \; t)
\end{align}
where $\mathbf{W}$ is a learnable parameter matrix. We note that the two edge types have different attention network parameters.

\par
For an event node $e_i$, let $\mathcal{N}_i^t$ denote the set of its neighbors each with an edge $e_{ji}^t$. That is, $e_j \in \mathcal{N}_i^t$ indicates a directed edge from $e_j$ to $e_i$. We next compute the weight $w_{ji}^t$ for the edge $e_{ji}$ as follows:
\begin{align}
w_{ji}^t = \frac{\exp(c_{ji}^t)}{\sum_{k \in \mathcal{N}_i^t}\exp(c_{ki}^t)}.
\end{align}
Now, we update the causal graph representation for the event node $e_i$ by the following multi-head attention mechanism as follows:
\begin{align}\label{Eq:NodeRepresentationUpdate}
	\mathbf{z}_i^t \leftarrow \parallel_{k=1}^K \sigma ( \sum_{j \in \mathcal{N}_i^t} w_{ji}^t \mathbf{W}^k\mathbf{z}_i )
\end{align}
where $\sigma(\cdot)$ denotes an element-wise activation function and $\parallel$ the concatenation operation.

\par
For an event $e_i$ with only one edge type, then its causal graph representation is updated by Equation~\eqref{Eq:NodeRepresentationUpdate}. For an event $e_i$ connected by two types of edges ($t_1$ for intra-sentence and $t_2$ for inter-sentence edges), we combine these two types of features with different weights to distinguish their confidence in ECI. Since two events within the same sentence are usually easier to identify, we tend to give more weight to the features related to intra-sentence edges. Finally, we combine the features $\mathbf{z}_i^{t_1}$ and $\mathbf{z}_i^{t_2}$ as follows:
\begin{align}
\mathbf{z}_i \leftarrow \beta \mathbf{z}_i^{t_1} + (1 - \beta) \mathbf{z}_i^{t_2}.
\end{align}
event pairs within the same sentence is relatively easier to identify

\subsection{Event Causality Identification}\label{Sec:ECI}
After learning the contextual text representation $\mathbf{h}$ and causal graph representation $\mathbf{z}$, we use a Multi-Layer Perceptron (MLP) to output a \textit{causal relation vector} $\mathbf{p}_{ij} \in \mathbb{R}^3$ for two events $e_i$ and $e_j$ as follows:
\begin{align}\label{Eq:ECI}
\mathbf{p}_{ij} = \mathsf{softmax}([\mathbf{h}_i\concat \mathbf{h}_j \concat (\mathbf{z}_i - \mathbf{z}_j)] \mathbf{W} ),
\end{align}
where $\concat$ stands for concatenation operation and $\mathbf{W}$ is a learnable parameter matrix. We note that the subtraction of $\mathbf{z}_i$ and $\mathbf{z}_j$ is to emphasize the causal directionality, as they are learned from a directed causal graph. We write $\mathbf{p}_{ij} =(p_{ij}^n, p_{ij}^c, p_{ij}^e)$ with the element $p_{ij}^n/p_{ij}^c/p_{ij}^e$ denoting the probability of a NONE/CAUSE/EFFECT relation between the two events.

\subsection{Causality Graph Construction}\label{Sec:CausalityGraphConstruction}
The CGC module is to construct a document-level \textit{event causality graph} (ECG) in each iteration. The ECG $\mathcal{G}=(\mathcal{N}, \mathcal{R})$ contains events as nodes, and event causal relations as edges. Considering the different information density~\citep{yuan2023discriminative} and recognition difficulty for inter-sentence event pairs and intra-sentence event pairs, we define two types of edges in $\mathcal{R}$: (1) Intra-sentence edges for two events in the same sentence, e.g., the green edge of \textit{rammed} $\rightarrow$ \textit{smashing} in Figure~\ref{fig:model}. (2) Inter-sentence edges for two events in two different sentences, e.g., the blue edge of \textit{drove} $\rightarrow$ \textit{stealing}  in Figure~\ref{fig:model}.

\par
To minimize error propagation, we resort to the causal relation vector $\mathbf{p}_{ij}$ by Equation~\eqref{Eq:ECI} to employ only those identified causal relations with high confidence as edges during iteration. We define $\omega$ as the relation confidence threshold. For two events $e_i$ and $e_j$, if $p_{ij}^c$ is the largest and $p_{ij}^c \ge \omega$, then a directed edge $e_{ij}$ is constructed from $e_i$ to $e_j$; If $p_{ij}^e$ is the largest and $p_{ij}^e \ge \omega$, then a directed edge $e_{ji}$ is constructed from $e_j$ to $e_i$. After processing all event pairs' $\mathbf{p}_{ij}$, we construct the ECG $\mathcal{G}$ with an \textit{adjacency matrix} $\mathbf{A}$: $\mathbf{A}_{ij}=1$, if there exists a directed edge from $e_i$ to $e_j$; Otherwise, $\mathbf{A}_{ij}=0$

\par
In our \textit{iterative Learning and Identifying Framework}, the ECG $\mathcal{G}$ is first initialized and then iteratively updated till the termination.

\par
\textbf{Initialization}: We initialize the causal graph representation for each event node as the event contextual text representation, i.e., $\mathbf{z}^{(0)}=\mathbf{h}$, and construct an ECG $\mathcal{G}^{(0)}$ based on the $\mathbf{z}^{(0)}$ and $\mathbf{h}$.

\par
\textbf{Iteration}: In the $l$-th iteration, we first use the previous ECG $\mathcal{G}^{(l-1)}$ to learn the new causal graph representations $\mathbf{z}^{(l)}$ for all nodes, as described in Section~\ref{Sec:CausalGraphEncoder}. We next construct a new ECG $\mathcal{G}^{(l)}$ based on the $\mathbf{z}^{(l)}$ and $\mathbf{h}$.

\par
\textbf{Termination}: To address the scale differences among documents and prevent over iteration, we design an iteration condition that terminates upon meeting either of the following two criteria. (1) The iteration count $l$ reaches a predefined \textit{maximum iteration number} $L$. For a document with $n<L$ sentences, we set its maximum number of iterations as $n$. (2) If the \textit{structural difference} between $\mathcal{G}^{(l)}$ and $\mathcal{G}^{(l-1)}$ is less than a predefined threshold:
\begin{align}\label{Eq:StructuralChangeFactor}
	\sum_i \sum_j | \mathbf{A}_{ij}^{(l)}-\mathbf{A}_{ij}^{(l-1)} | \le \delta.
\end{align}
After terminating the iteration, all the directed edges in the final ECG $\mathcal{G}$ are the identified document-level causal relations.

\subsection{Training strategy}\label{Sec:TrainingStrategy}
Intuitively, the ECG $\mathcal{G}^{(l)}$ in each iteration carries different significance for the final ECG. We differentiate the iterations by first calculating an iteration-level loss $\mathcal{L}_l$ for the $l$-th iteration. We adopt the $\alpha$-balanced variant of focal loss to address the issue of class imbalance and compute  $\mathcal{L}_l$ by
\begin{align}
	\mathcal{L}_l=-\sum_{e_i,e_j \in \mathcal{D}} \alpha^c (1-\hat{p}_{ij}^c)^\gamma \mathrm{log}(\hat{p}_{ij}^c)
 	\label{Eq:Loss},
\end{align}
where $\hat{p}_{ij}^c $ is the predicted probability of the true class $c$, $\alpha^c$ is a weighting factor for the true class $c$. Additionally, $\gamma$ represents a predefined focusing hyper-parameter.

\par
We note that during the  identifying while learning process, if some misidentification happens in an iteration, it may propagate to the later iterations. To penalize such propagations, we emphasize the importance of earlier iterations and introduce a balancing factor inversely proportional to the iteration count in the final loss function. We define the final loss function as follows:
\begin{align}\label{Eq:FinalLoss}
	\mathcal{L}=\sum_{l=1}^{L_\mathcal{D}} \frac{1}{l} \mathcal{L}_l,
\end{align}
where $L_\mathcal{D}$ denotes the actual iteration counts of the document $\mathcal{D}$.
\begin{table*}[h]
	\centering
	\resizebox{0.9 \linewidth}{!}{
			\renewcommand\arraystretch{1}
			\begin{tabular}{l|ccc|ccc|ccc}
				\toprule
				\multirow{2}{*}{\textbf{Model}} & \multicolumn{3}{c|}{\text{Intra-sentence}} & \multicolumn{3}{c|}{\text{Inter-sentence}} & \multicolumn{3}{c}{\text{Intra+Inter}}  \\
				\cmidrule(lr){2-4} \cmidrule(lr){5-7} \cmidrule(lr){8-10} & P(\%)     & R(\%)     & F1(\%)     & P(\%)     & R(\%)     & F1(\%)     & P(\%)   & R(\%)   & F1(\%)  \\ \midrule
				\multicolumn{10}{c}{\textit{Causality Direction Identification}} \\ \midrule
				\text{BERT} \citep{devlin2018bert}      & \underline{62.4}       & 32.6      & 42.8              & 34.4     & 30.7    & 32.4             & 40.7     & 31.3     & 35.4    \\
				\text{RoBERTa} \citep{liu2019roberta}  & 59.7       & 38      & 46.4              & 31.3     & 34.2    & 32.7             & 37.3     & 35.5     & 36.4    \\
				\text{LONG} \citep{beltagy2020longformer}    & 59.0          & 40.5         & 48.0              & 35.2        & 30.5       & 32.7  & 41.6        & 33.8        & 37.3    \\
				\text{ERGO} \citep{chen2022ergo}     & 58.8     & 47.6   & 52.6  & 36.1   & \underline{41.2}    & 38.5            & 41.5     & 43.3     & 42.4    \\
				\text{SENDIR} \citep{yuan2023discriminative}     & 56.0     & \underline{52.6}    & \underline{54.2}  & \underline{38.6}   & 39.4    & \underline{39.0}             & \underline{43.8}     & \underline{43.7}     & \underline{43.7}     \\
				\text{LLaMA-2-7B} \citep{gao2023chatgpt}     & 17.5     & 17    & 17.2  & 6.8   & 19.2    & 10.0             & 8.3     & 18.5     & 11.5      \\
				\midrule
				\textbf{iLIF} (the proposed)      & \textbf{66.7}     & \textbf{54.5}    & \textbf{60.0}     & \textbf{41.2}   & \textbf{44.6}  & \textbf{42.8}    & \textbf{47.9}   & \textbf{47.8}   & \textbf{47.8}    \\ \midrule
				\multicolumn{10}{c}{\textit{Causality Existence Identification}} \\ \midrule
				\text{BERT} \citep{devlin2018bert}      & 60.4       & 45.7      & 52.0              & 30.6     & 39.1    & 34.3             & 37.2     & 41.2     & 39.1    \\
				\text{RoBERTa} \citep{liu2019roberta}  & 62.7       & 45.4      & 52.7              & 32.7     & 38.3    & 35.3             & 39.7     & 40.6     & 40.1    \\
				\text{LONG} \citep{beltagy2020longformer}    & 47.7          & 69.3         & 56.5              & 26.1        & 55.6       & 35.5  & 31.4        & 60.0        & 41.2    \\
				\text{ERGO} \citep{chen2022ergo}     & 49.7     & 72.6   & 59.0  & \underline{43.2}   & 48.8    & 45.8            & \underline{46.3}     & 50.1     & 48.1    \\
				\text{SENDIR} \citep{yuan2023discriminative}     & \underline{65.8}     & 66.7    & \underline{66.2}  & 33.0   & \textbf{90.0}    & \underline{48.3}             & 37.8     & \textbf{82.8}     & \underline{51.9}     \\
				\text{text-davinci-003} \citep{gao2023chatgpt}     & 33.2     & 74.4    & 45.9  & -   & -    & -             & -     & -     & -      \\
				\text{gpt-3.5-turbo} \citep{gao2023chatgpt}     & 27.6     & \underline{80.2}    & 41.0  & -   & -    & -             & -     & -     & -      \\
				\text{gpt-4} \citep{gao2023chatgpt}     & 27.2     & \textbf{94.7}    & 42.2  & -   & -    & -             & -     & -     & -      \\
				\text{LLaMA-2-7B} \citep{gao2023chatgpt}     & 26.9     & 29.3    & 28.0  & 10.8   & 31.9    & 16.1             & 13.2     & 31.1     & 18.5     \\
				\midrule
				\textbf{iLIF} (the proposed)     & \textbf{76.8}     & 66.3    & \textbf{71.2}     & \textbf{53.5}   & \underline{65.9}  & \textbf{59.1}    & \textbf{59.2}   & \underline{66.1}   & \textbf{62.5}    \\
				\bottomrule
		\end{tabular}}
		\caption{Overall results on the EventStoryLine dataset in both direction and existence evaluation settings: The best results are highlighted in \textbf{bold}, and the second-best results are \underline{underlined}. Intra-sentence indicates that the event pair is within the same sentence, while Inter-sentence indicates that the event pair is in different sentences.}
		\label{Tab:ESC}
	\end{table*}

\begin{table*}[h]
	\centering
	\resizebox{0.9\linewidth}{!}{
			\renewcommand\arraystretch{1}
			\begin{tabular}{l|ccc|ccc|ccc}
				\toprule
				\multirow{2}{*}{\textbf{Model}} & \multicolumn{3}{c|}{\text{Intra-sentence}} & \multicolumn{3}{c|}{\text{Inter-sentence}} & \multicolumn{3}{c}{\text{Intra+Inter}}  \\
				\cmidrule(lr){2-4} \cmidrule(lr){5-7} \cmidrule(lr){8-10} & P(\%)     & R(\%)     & F1(\%)     & P(\%)     & R(\%)     & F1(\%)     & P(\%)   & R(\%)   & F1(\%)  \\ \midrule
				\multicolumn{10}{c}{\textit{Causality Direction Identification}} \\ \midrule
				\text{BERT} \citep{devlin2018bert}      & 46.0       & 48.8      & 47.4              & 42.1     & 45.2    & 43.6             & 42.4     & 45.5     & 43.9    \\
				\text{ERGO} \citep{chen2022ergo}     & \underline{62.3}     & \textbf{63.1}   &\textbf{62.7}  & \underline{47.8}   & \textbf{59.8}    & \underline{53.1}            & \underline{48.7}     & \textbf{60.1}     & \underline{53.8}    \\
				\text{SENDIR} \citep{yuan2023discriminative}     & 46.6      & 44.2     & 45.4   & 46.8    & 43.0     & 44.8    & 46.8   & 43.1   & 44.9    \\
				\midrule
				\textbf{iLIF} (the proposed)       & \textbf{73.7}     & \underline{49.7}    & \underline{59.4}     & \textbf{66.3}   & \underline{47.5}  & \textbf{55.3}    & \textbf{66.9}   & \underline{47.6}   & \textbf{55.6}    \\ \midrule
				\multicolumn{10}{c}{\textit{Causality Existence Identification}} \\ \midrule
				\text{BERT} \citep{devlin2018bert}      & 46.8      & 50.3      & 48.5              & 43.0     & 46.8    & 44.8             & 43.3     & 47.1     & 45.1    \\
				\text{ERGO} \citep{chen2022ergo}     & \underline{63.1}     & 65.3   & \textbf{64.2}   & 48.7   & \textbf{62.0}    & \underline{54.6}            & 49.6     & \textbf{62.3}     & \underline{55.2}    \\
				\text{SENDIR} \citep{yuan2023discriminative}     & 51.4      & 53.6     & 52.5    & \underline{51.9}    & \underline{52.8}     & 52.4    & \underline{51.9}   & \underline{52.9}   & 52.4    \\
				\text{text-davinci-003} \citep{gao2023chatgpt}     & 25.0     & 75.1    & 37.5  & -   & -    & -             & -     & -     & -      \\
				\text{gpt-3.5-turbo} \citep{gao2023chatgpt}     & 19.9     & \underline{85.8}    & 32.3 & -   & -    & -             & -     & -     & -      \\
				\text{gpt-4} \citep{gao2023chatgpt}     & 22.5     & \textbf{92.4}    & 36.2  & -   & -    & -             & -     & -     & -      \\
				\midrule
				\textbf{iLIF} (the proposed)       & \textbf{74.4}     & 51.5    & \underline{60.9}    & \textbf{67.1}   & 49.2   & \textbf{56.8}    & \textbf{67.7}	& 49.4	& \textbf{57.1}\\
				\bottomrule
		\end{tabular}}
		\caption{Overall results on the MAVEN-ERE dataset in both direction and existence evaluation settings.}
		\label{Tab:MAVEN-ERE}
	\end{table*}

\section{Experiments}\label{Sec:Experiments}
\subsection{Experimental Settings}\label{Sec:DatasetsAndEvaluationMetrics}
\paragraph{Datasets Details} We evaluate our \textsf{iLIF} on the widely used EventStoryLine (v0.9) dataset and MAVEN-ERE dataset. The EventStoryLine dataset comprises 22 topics, 258 documents, 5,334 event mentions, 1,770 intra-sentence causal event pairs and 3,855 inter-sentence causal event pairs~\citep{caselli2017event}. Following~\citep{caselli2017event, gao2019modeling}, we designate event pairs annotated with `PRECONDITION' as CAUSE relation, and  `FALLING$\_$ACTION' as EFFECT relation. In both settings, we utilize the last two topics as the development dataset, leaving the remaining 20 topics for 5-fold cross-validation.
\par
The MAVEN-ERE dataset is a large-scale dataset, which comprises 4,480 documents, 103,193 events, and 57,992 causal event pairs. These causal event pairs are annotated as `CAUSE' or `PRECONDITION,' both representing CAUSE relations. We randomly reverse half of the event pairs with a CAUSE relation to represent the EFFECT relation. As MAVEN-ERE did not release the test set, following \citet{tao2023seag}, we use the original development set as the test set. Additionally, we sample 10$\%$ of the data from the original training set to form the development set.
\par
\paragraph{implementation details} Our method is implemented based on the PyTorch version of Huggingface Transformer~\citep{wolf2020transformers}. We use the uncased BERT-base~\citep{devlin2018bert} as the base PLM and fine-tune it during the training process. We optimize our model using AdamW~\citep{loshchilov2018decoupled}, with a linear warm-up for the first 10$\%$ of steps. The learning rate for the PLM is set to 2e-5, while for other modules, it is set to 1e-4. The batch size is set to 1. More details can be found in Appendix~\ref{Sec:DetailsAboutexperimentalSettings}.
\par
\paragraph{Evaluation Metrics} We adopt the commonly used Precision ($\mathsf{P}$), Recall ($\mathsf{R}$), and F1-score ($\mathsf{F}1$) as the evaluation metrics. In the direction evaluation, we calculate the micro-averaged results for Precision, Recall, and F1-score specifically for the CAUSE and EFFECT classes. In the existence evaluation, for the EventStoryLine dataset, we follow the same approach as previous methods to ensure fair comparison~\citep{phu2021graph}.
\par
\paragraph{Direction and Existence Settings}
In the direction setting, we utilize three labels: {NONE, CAUSE, EFFECT}, which respectively represent the noncausal relation, the cause relation, and the effect relation. The \textit{direction identification} results of competitors are derived by expanding the classification module of the competitors' algorithms from binary to ternary classification. In the existence setting, we adopt two labels: {NONE, CAUSAL}, which respectively represent noncausal relation and causal relation. For MAVEN-ERE dataset, we combine the event pairs identified as the CAUSE class and the EFFECT class in the direction experiment into the CAUSAL class event pairs to obtain the \textit{existence identification} result.

\subsection{Competitors}\label{Sec:Baselines}
We compare \textsf{iLIF} with the following competitors:

\par
\textsf{PLM-base} concatenates two events' contextual text representations and then identifies causality relation using a MLP. We use BERT-base~\citep{devlin2018bert}, RoBERTa-base~\citep{liu2019roberta}, and Longformer-base~\citep{beltagy2020longformer} as the PLM.

\par
\textsf{ERGO}~\citep{chen2022ergo} builds a relational graph to model interactions between event pairs.

\par
\textsf{SENDIR}~\citep{yuan2023discriminative} leverages intra-sentence event pairs to construct a reasoning chain, facilitating inter-sentence causality reasoning.

\par
\textsf{Large Language Models (LLMs).} \citet{gao2023chatgpt} conduct zero-shot ECI experiments using OpenAI’s official API~\footnote{\url{https://platform.openai.com/}}, covering three versions of ChatGPT: text-davinci-003, gpt-3.5-turbo and gpt-4. We also test another popular LLM, the LLaMA2~\citep{touvron2023llama} of Llama-2-7b-chat version. Appendix~\ref{app:LLaMA2PromptDetails} reports the designed prompts for the LLMs.

\par
We consider both \textit{existence identification} and \textit{direction identification} for performance evaluations. The existence identification means to only identify the existence of causal relation between two events with causality direction. The direction identification means to correctly identify the causal direction, if existing, between two events.

\begin{table}[t]
	\centering
	\resizebox{\columnwidth}{!}{
			\renewcommand\arraystretch{1.2}
			\begin{tabular}{l|ccc|ccc}
				\toprule
				\multirow{3}{*}{\textbf{Model}} & \multicolumn{3}{c|}{\textbf{Direction}} & \multicolumn{3}{c}{\textbf{Existence}} \\
				\cmidrule(lr){2-4} \cmidrule(lr){5-7} & P(\%)     & R(\%)     & F1(\%)     & P(\%)     & R(\%)     & F1(\%)   \\ \midrule
				\textbf{iLIF}     & 47.9   & 47.8   & 47.8  & 59.2   & 66.1   & 62.5 \\
				\textbf{iLIF} w/o \text{Direction}   & 42.3   & 49.3   & 45.5  &	55.9	&	63.8	& 59.6 \\
				\textbf{iLIF} w/o \text{Heterogeneity}   & 45.5   & 47.2   & 46.3  &	56.4	&	66.3	& 61.0   \\
				\textbf{iLIF} w/o \text{Iteration}     & 43.2   & 45.2   & 44.2   & 52.8   & 65.9   & 58.6 \\
				\bottomrule
		\end{tabular}}
		\caption{Results of ablation study on EventStoryLine.}
		\label{Tab:Ablation}
	\end{table}

\subsection{Overall Results}
Table~\ref{Tab:ESC} and Table~\ref{Tab:MAVEN-ERE} compare the overall results on the EventStoryLine and MAVEN-ERE dataset, respectively. We note that all the competitors adopt the \textit{identifying after learning} paradigm; While our \textsf{iLIF} adopts the \textit{identifying while learning}. Our \textsf{iLIF} achieves the best overall F1 results (intra+inter) on the two datasets in both existence and direction evaluation. This validates the superiority of our \textsf{iLIF} model with the new yet effective identifying while learning mode for the ECI task. For those LLMs, although they are capable of zero-shot causal reasoning, their performance is much worse than those methods based on the fine-tuned small PLMs. This suggests that LLMs, like ChatGPT and LLAMA, may not be effective causal reasoners for complex causal reasoning tasks.

\par
Taking a close observation of the two tables, all models perform better on identifying causality in intra-sentence than that in inter-sentence. This is in accordance with the report in ~\citet{yuan2023discriminative}, as comprehending events in one sentence is generally easier with the same sentential context. We note that our \textsf{iLIF} achieves large improvements on the Precision in the intra-sentence causality identification. This can be attributed to our using a high confidence threshold, by which the intra-sentence causal relations are often with high identification confidence; While this also leads to a relatively lower Recall compared with other models. However, using such identified intra-sentence relations with high confidence can help improving inter-sentence identification, as the constructed event causality graph is becoming more confident with the iterations, on which events' causal structure representations can be well learned to further boost inter-sentence causality identification. This is evidenced from the high Precision and F1-score in the inter-sentence identification, and as a result, the overall intra+inter identification of our \textsf{iLIF} performs better than the competitors.

\begin{figure}[t]
	\centering
	\includegraphics[width=0.95\columnwidth,height=0.32\textwidth]{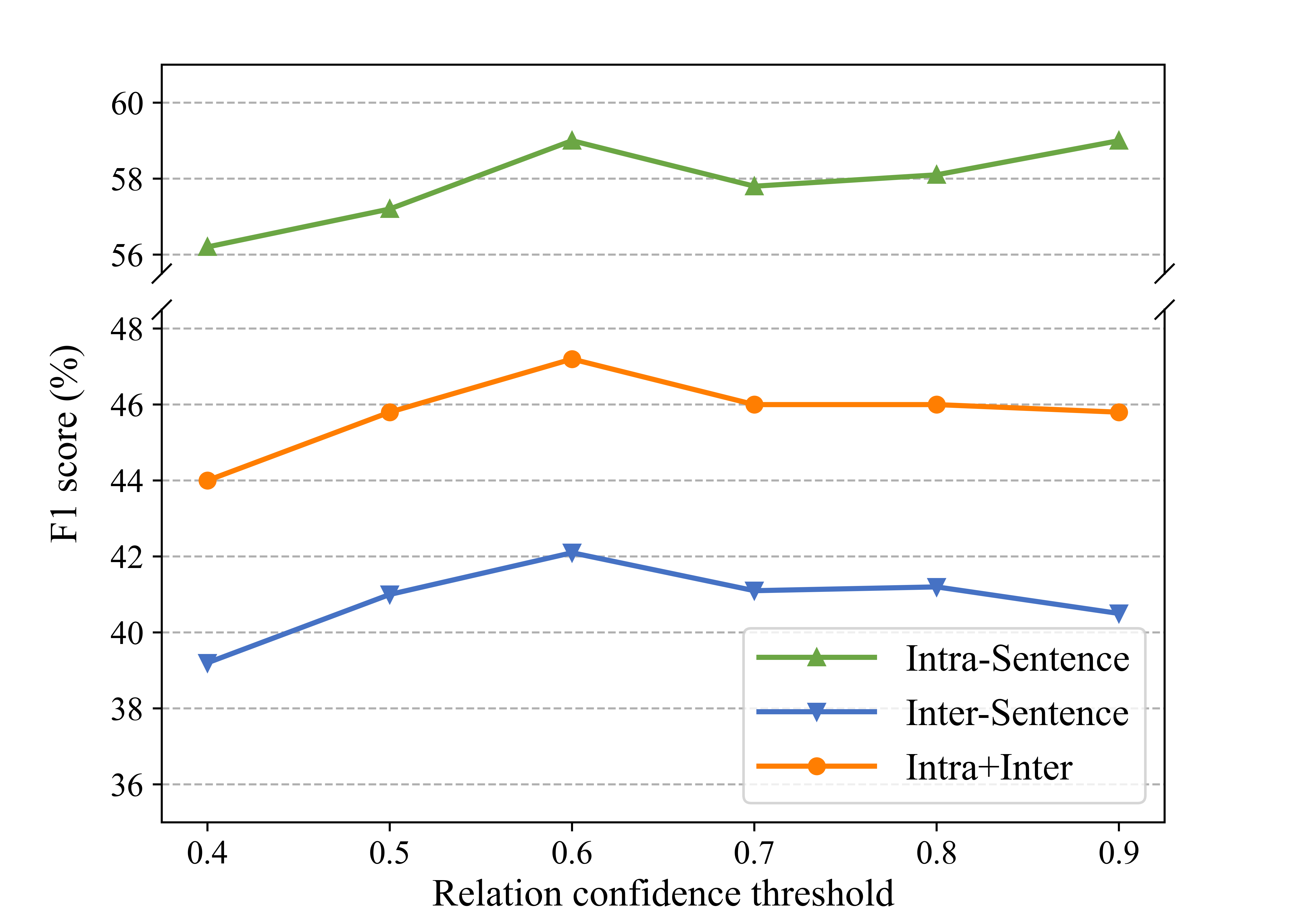}
	\caption{F1 scores on EventStoryLine when using different edge thresholds in the direction setting.}
	\label{Fig:Threshold}
\end{figure}

\begin{figure}[t]
	\centering
	\includegraphics[width=0.95\columnwidth,height=0.35\textwidth]{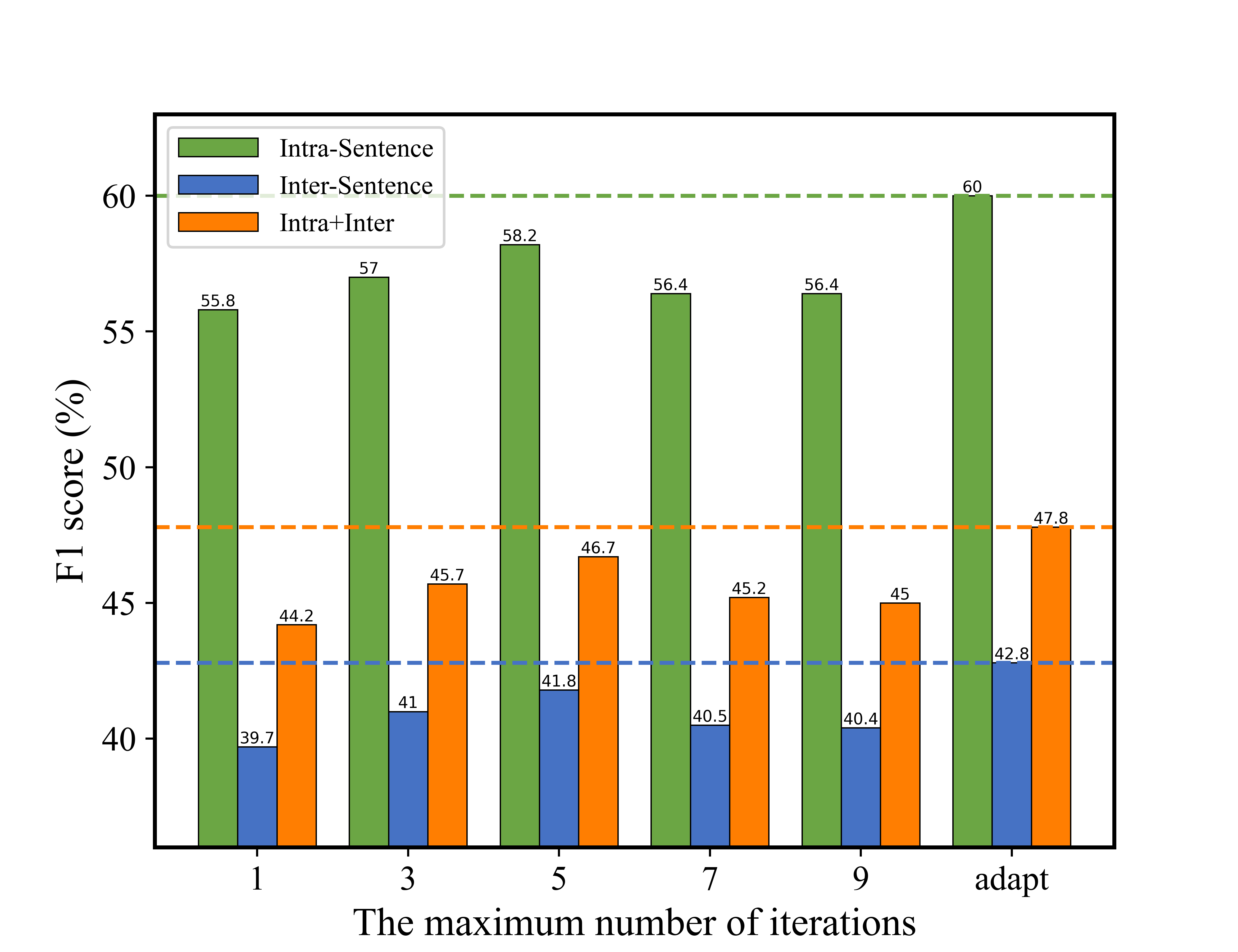}
	\caption{Results of different maximum number of iterations on EventStoryLine in the direction setting.}
	\label{Fig:Iteration}
\end{figure}

\subsection{Ablation Study}
Table~\ref{Tab:Ablation} presents the ablation studies for examining the module functionalities.

\par
(1) \textsf{iLIF w/o Direction}, which removes the directionality of the ECG. The removal of directionality causes a decrease in F1-scores by 4.8\% and 4.6\% in the two evaluations, respectively. Additionally, the decrease is primarily concentrated in the Precision. These results imply the importance of encoding causal direction and structure information on a directed ECG for learning potential events' interactions. We also observe that removing the direction increases the Recall in the direction evaluation. One possible reason is that the absence of direction increases the number of edges for encoding more interaction information. However, this also introduces more noisy edges, leading to inaccurate causal structure representations and worse F1 scores.

\par
(2) \textsf{iLIF w/o Heterogeneity}, which only set one type of edges in the ECG, without distinguishing the intra-sentence and inter-sentence edges. We observe that removing edge heterogeneity results in a reduction in the Precision and F1-scores. One possible reason is that recognition difficulty does differ in the intra- and inter-sentence cases; While neglecting such differences cannot well utilize more confident intra-sentence relations when encoding on the ECG.
\par
(3) \textsf{iLIF w/o Iteration}, which constructs the ECG only once for learning causal graph representations to output final identifications. As it does not evolve with the re-identified causal relations, it cannot enjoy the potentials of learning more accurate events' causal graph representations in the later iterations, so resulting in decreases of identification performance. The results again validate the effectiveness of our \textit{identifying while learning} mode.


\subsection{Parameter Study}\label{sec:EdgeThresholdAnalysis}
\paragraph{Relation Confidence Threshold}
Figure~\ref{Fig:Threshold} presents the results on examining the impact of using different relation confidence thresholds. It can be observed that using too low or two high thresholds introduces performance decrease. This is not unexpected. Using too low threshold would introduce more edges in the ECG, yet some may be incorrect ones, leading to inaccurate causal graph representation learning on the ECG. Setting the threshold too high results in a sparse ECG, which restricts the quality of the learned causal graph representation. Experiment results suggest to set 0.6 as the relation confidence threshold.
\begin{figure}[t]
	\centering
	\includegraphics[width=0.38\textwidth, height=0.32\textwidth]{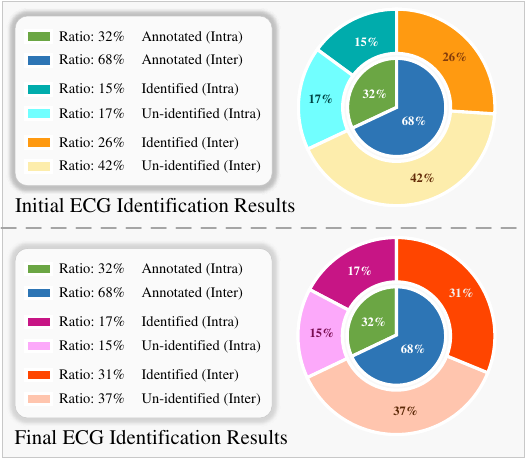}
	\caption{Comparison of causality identification results in the initial ECG and final ECG.  }
	\label{Fig:ECGInitialFinal}
\end{figure}

\begin{figure*}[h]
	\centering
	\includegraphics[width=0.92\textwidth, height=0.28\textwidth]{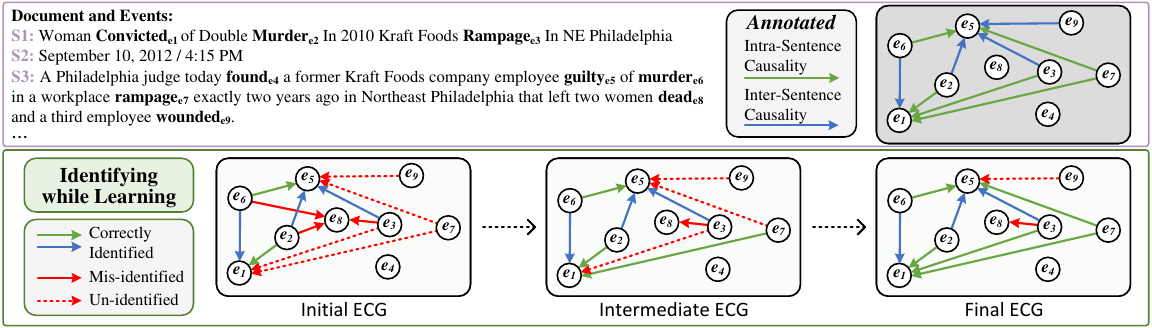}
	\caption{A case of the identifying while learning process on the EventStoryLine dataset. The initial ECG is with 3 mis-identified and 4 un-identified causal relations. After learning and updating events' causal graph representations from a few ECGs, some mis-identified relations can be corrected and some un-identified relations can be identified. At the termination, the final ECG is with only 1 mis-identified and 1 un-identified causal relation. }
	\label{Fig:Case}
\end{figure*}
\paragraph{Number of Iterations}
Figure~\ref{Fig:Iteration} presents the results on examining using different numbers of iterations. We can fix the number of iterations or use our termination criterion to adaptively adjust maximum iterations for different documents. As different documents may contain different numbers of sentences, it can be observed that using such fixed setting cannot well adapt to the length of documents. Furthermore, our criterion also enables to terminate iterations for two consecutive ECGs with small topological difference, hence helping to balance the under-fitting or over-fitting issue in the causal representation learning process.
\par
Increasing the number of iterations not only affects the results but also increases the time complexity of the algorithm. We compare the complexity of different algorithms from both temporal and spatial perspectives. Experimental details can be found in Appendix~\ref{app:AlgorithmComplexityComparison}. The results demonstrate that our algorithm achieves a trade-off between algorithmic complexity and identification performance.

\subsection{Exploration Study}
Figure~\ref{Fig:ECGInitialFinal} compares the identification results of the initial and final ECG on the EventStoryLine testing dataset, which contains $32\%$ intra-sentence and $68\%$ inter-sentence causal relations. Compared with the initial ECG, the final ECG can increase the ratio of correct identification and decrease the ratio of un-identification for both intra-sentence and inter-sentence causal relations. This indicates that causality identification can benefit from the updating of events' representations during the identifying while learning process. Figure~\ref{Fig:Case} presents a case of the identifying while learning process: Compared with the annotated intra-sentence and inter-sentence causality (i.e., the ground truth), the initially constructed ECG is with 3 mis-identified and 4 un-identified causal relations; While after learning and updating events' causal graph representations from a few ECGs, 2 mis-identified relations can be corrected and 1 un-identified can be correctly identified in an intermediate ECG. At the termination, the final ECG reduces mis-identified relations from 3 to 1 and un-identified from 4 to 1.


\section{Conclusion}
In this paper, we propose a novel \textit{identifying while learning} mode where the central idea is to iteratively update events' representations for boosting next round of causality identification. Within the proposed \textit{iterative Learning and Identifying Framework}, an event causality graph is constructed in each iteration based on previously identified causal relations with high confidence, which helps to mine events' directed interactions for updating their representations. Experiments on two widely used datasets have validated the superiority of this new working mode.

\section*{Acknowledgements}
This work is supported in part by National Natural Science Foundation of China (Grant No:62172167). The computation is supported by the HPC Platform of Huazhong University of Science and Technology.

\section*{Limitation}
We preprocess the EventStoryLine dataset~\citep{caselli2017event} to ensure that each ground truth ECG is a directed acyclic graph~\citep{gopnik2007causal}. However, We do not guarantee that the final \textit{event causality graph} is a directed acyclic graph. In future work, we plan to introduce the directed acyclic constraint into the causality graph construction process to enhance the practical application effectiveness of the model.

\section*{Ethics Statement}
Our research work meets the ethics of ACL. The proposed identifying while learning model can identify causal relations among events in a document. However, the algorithm is not perfect and may result in erroneous predictions. Therefore, researchers should not rely solely on the model to make real-world decisions.


\bibliography{anthology,custom}

\begin{thebibliography}{38}
\expandafter\ifx\csname natexlab\endcsname\relax\def\natexlab#1{#1}\fi

\bibitem[{Al-Khatib et~al.(2020)Al-Khatib, Hou, Wachsmuth, Jochim, Bonin, and
  Stein}]{al2020end}
Khalid Al-Khatib, Yufang Hou, Henning Wachsmuth, Charles Jochim, Francesca
  Bonin, and Benno Stein. 2020.
\newblock End-to-end argumentation knowledge graph construction.
\newblock In \emph{Proceedings of the AAAI conference on artificial
  intelligence}, volume~34, pages 7367--7374.

\bibitem[{Banarescu et~al.(2013)Banarescu, Bonial, Cai, Georgescu, Griffitt,
  Hermjakob, Knight, Koehn, Palmer, and Schneider}]{banarescu2013abstract}
Laura Banarescu, Claire Bonial, Shu Cai, Madalina Georgescu, Kira Griffitt, Ulf
  Hermjakob, Kevin Knight, Philipp Koehn, Martha Palmer, and Nathan Schneider.
  2013.
\newblock Abstract meaning representation for sembanking.
\newblock In \emph{Proceedings of the 7th linguistic annotation workshop and
  interoperability with discourse}, pages 178--186.

\bibitem[{Beltagy et~al.(2020)Beltagy, Peters, and
  Cohan}]{beltagy2020longformer}
Iz~Beltagy, Matthew~E Peters, and Arman Cohan. 2020.
\newblock Longformer: The long-document transformer.
\newblock \emph{arXiv preprint arXiv:2004.05150}.

\bibitem[{Cao et~al.(2021)Cao, Zuo, Chen, Liu, Zhao, Chen, and
  Peng}]{cao2021knowledge}
Pengfei Cao, Xinyu Zuo, Yubo Chen, Kang Liu, Jun Zhao, Yuguang Chen, and Weihua
  Peng. 2021.
\newblock Knowledge-enriched event causality identification via latent
  structure induction networks.
\newblock In \emph{Proceedings of the 59th Annual Meeting of the Association
  for Computational Linguistics and the 11th International Joint Conference on
  Natural Language Processing (Volume 1: Long Papers)}, pages 4862--4872.

\bibitem[{Caselli and Vossen(2017)}]{caselli2017event}
Tommaso Caselli and Piek Vossen. 2017.
\newblock The event storyline corpus: A new benchmark for causal and temporal
  relation extraction.
\newblock In \emph{Proceedings of the Events and Stories in the News Workshop},
  pages 77--86.

\bibitem[{Chen et~al.(2022)Chen, Cao, Deng, Li, Wang, Shao, and
  Zhang}]{chen2022ergo}
Meiqi Chen, Yixin Cao, Kunquan Deng, Mukai Li, Kun Wang, Jing Shao, and Yan
  Zhang. 2022.
\newblock Ergo: Event relational graph transformer for document-level event
  causality identification.
\newblock \emph{arXiv preprint arXiv:2204.07434}.

\bibitem[{Chen et~al.(2023)Chen, Cao, Zhang, and Liu}]{chen2023cheer}
Meiqi Chen, Yixin Cao, Yan Zhang, and Zhiwei Liu. 2023.
\newblock Cheer: Centrality-aware high-order event reasoning network for
  document-level event causality identification.

\bibitem[{Chen et~al.(2019)Chen, Chang, Chen, Nayak, and Ku}]{chen2019uhop}
Zi-Yuan Chen, Chih-Hung Chang, Yi-Pei Chen, Jijnasa Nayak, and Lun-Wei Ku.
  2019.
\newblock Uhop: An unrestricted-hop relation extraction framework for
  knowledge-based question answering.
\newblock \emph{arXiv preprint arXiv:1904.01246}.

\bibitem[{Devlin et~al.(2018)Devlin, Chang, Lee, and
  Toutanova}]{devlin2018bert}
Jacob Devlin, Ming-Wei Chang, Kenton Lee, and Kristina Toutanova. 2018.
\newblock Bert: Pre-training of deep bidirectional transformers for language
  understanding.
\newblock \emph{arXiv preprint arXiv:1810.04805}.

\bibitem[{Do et~al.(2011)Do, Chan, and Roth}]{do2011minimally}
Quang Do, Yee~Seng Chan, and Dan Roth. 2011.
\newblock Minimally supervised event causality identification.
\newblock In \emph{Proceedings of the 2011 conference on empirical methods in
  natural language processing}, pages 294--303.

\bibitem[{Fan et~al.(2022)Fan, Liu, Qin, Zhang, and Xu}]{fan2022towards}
Chuang Fan, Daoxing Liu, Libo Qin, Yue Zhang, and Ruifeng Xu. 2022.
\newblock Towards event-level causal relation identification.
\newblock In \emph{Proceedings of the 45th International ACM SIGIR Conference
  on Research and Development in Information Retrieval}, pages 1828--1833.

\bibitem[{Fellbaum(1998)}]{fellbaum1998wordnet}
Christiane Fellbaum. 1998.
\newblock \emph{WordNet: An electronic lexical database}.
\newblock MIT press.

\bibitem[{Gao et~al.(2023)Gao, Ding, Qin, and Liu}]{gao2023chatgpt}
Jinglong Gao, Xiao Ding, Bing Qin, and Ting Liu. 2023.
\newblock Is chatgpt a good causal reasoner? a comprehensive evaluation.
\newblock \emph{arXiv preprint arXiv:2305.07375}.

\bibitem[{Gao et~al.(2019)Gao, Choubey, and Huang}]{gao2019modeling}
Lei Gao, Prafulla~Kumar Choubey, and Ruihong Huang. 2019.
\newblock Modeling document-level causal structures for event causal relation
  identification.
\newblock In \emph{Proceedings of the 2019 Conference of the North American
  Chapter of the Association for Computational Linguistics: Human Language
  Technologies, Volume 1 (Long and Short Papers)}, pages 1808--1817.

\bibitem[{Gopnik et~al.(2007)Gopnik, Schulz, and Schulz}]{gopnik2007causal}
Alison Gopnik, Laura Schulz, and Laura~Elizabeth Schulz. 2007.
\newblock \emph{Causal learning: Psychology, philosophy, and computation}.
\newblock Oxford University Press.

\bibitem[{He et~al.(2021)He, Cui, Shen, Xu, Liu, and Jiang}]{he2021daring}
Yue He, Peng Cui, Zheyan Shen, Renzhe Xu, Furui Liu, and Yong Jiang. 2021.
\newblock Daring: Differentiable causal discovery with residual independence.
\newblock In \emph{Proceedings of the 27th ACM SIGKDD Conference on Knowledge
  Discovery \& Data Mining}, pages 596--605.

\bibitem[{Hu et~al.(2023)Hu, Li, Jin, Bai, Guan, Guo, and
  Cheng}]{hu2023semantic}
Zhilei Hu, Zixuan Li, Xiaolong Jin, Long Bai, Saiping Guan, Jiafeng Guo, and
  Xueqi Cheng. 2023.
\newblock Semantic structure enhanced event causality identification.
\newblock \emph{arXiv preprint arXiv:2305.12792}.

\bibitem[{Liu et~al.(2021)Liu, Chen, and Zhao}]{liu2021knowledge}
Jian Liu, Yubo Chen, and Jun Zhao. 2021.
\newblock Knowledge enhanced event causality identification with mention
  masking generalizations.
\newblock In \emph{Proceedings of the Twenty-Ninth International Conference on
  International Joint Conferences on Artificial Intelligence}, pages
  3608--3614.

\bibitem[{Liu et~al.(2019)Liu, Ott, Goyal, Du, Joshi, Chen, Levy, Lewis,
  Zettlemoyer, and Stoyanov}]{liu2019roberta}
Yinhan Liu, Myle Ott, Naman Goyal, Jingfei Du, Mandar Joshi, Danqi Chen, Omer
  Levy, Mike Lewis, Luke Zettlemoyer, and Veselin Stoyanov. 2019.
\newblock Roberta: A robustly optimized bert pretraining approach.
\newblock \emph{arXiv preprint arXiv:1907.11692}.

\bibitem[{Loshchilov and Hutter(2018)}]{loshchilov2018decoupled}
Ilya Loshchilov and Frank Hutter. 2018.
\newblock Decoupled weight decay regularization.
\newblock In \emph{International Conference on Learning Representations}.

\bibitem[{Oh et~al.(2017)Oh, Torisawa, Kruengkrai, Iida, and
  Kloetzer}]{oh2017multi}
Jong-Hoon Oh, Kentaro Torisawa, Canasai Kruengkrai, Ryu Iida, and Julien
  Kloetzer. 2017.
\newblock Multi-column convolutional neural networks with causality-attention
  for why-question answering.
\newblock In \emph{Proceedings of the Tenth ACM international conference on web
  search and data mining}, pages 415--424.

\bibitem[{Pearl and Mackenzie(2018)}]{pearl2018book}
Judea Pearl and Dana Mackenzie. 2018.
\newblock \emph{The book of why: the new science of cause and effect}.
\newblock Basic books.

\bibitem[{Phu and Nguyen(2021)}]{phu2021graph}
Minh~Tran Phu and Thien~Huu Nguyen. 2021.
\newblock Graph convolutional networks for event causality identification with
  rich document-level structures.
\newblock In \emph{Proceedings of the 2021 conference of the North American
  chapter of the association for computational linguistics: Human language
  technologies}, pages 3480--3490.

\bibitem[{Pitler et~al.(2009)Pitler, Louis, and Nenkova}]{pitler2009automatic}
Emily Pitler, Annie Louis, and Ani Nenkova. 2009.
\newblock Automatic sense prediction for implicit discourse relations in text.

\bibitem[{Schuler(2005)}]{schuler2005verbnet}
Karin~Kipper Schuler. 2005.
\newblock \emph{VerbNet: A broad-coverage, comprehensive verb lexicon}.
\newblock University of Pennsylvania.

\bibitem[{Shen et~al.(2022)Shen, Zhou, Wu, and Qi}]{shen2022event}
Shirong Shen, Heng Zhou, Tongtong Wu, and Guilin Qi. 2022.
\newblock Event causality identification via derivative prompt joint learning.
\newblock In \emph{Proceedings of the 29th International Conference on
  Computational Linguistics}, pages 2288--2299.

\bibitem[{Tao et~al.(2023)Tao, Jin, Bai, Zhao, Dou, Zhao, Wang, and
  Tao}]{tao2023seag}
Zhengwei Tao, Zhi Jin, Xiaoying Bai, Haiyan Zhao, Chengfeng Dou, Yongqiang
  Zhao, Fang Wang, and Chongyang Tao. 2023.
\newblock Seag: Structure-aware event causality generation.
\newblock In \emph{Findings of the Association for Computational Linguistics:
  ACL 2023}, pages 4631--4644.

\bibitem[{Touvron et~al.(2023)Touvron, Lavril, Izacard, Martinet, Lachaux,
  Lacroix, Rozi{\`e}re, Goyal, Hambro, Azhar et~al.}]{touvron2023llama}
Hugo Touvron, Thibaut Lavril, Gautier Izacard, Xavier Martinet, Marie-Anne
  Lachaux, Timoth{\'e}e Lacroix, Baptiste Rozi{\`e}re, Naman Goyal, Eric
  Hambro, Faisal Azhar, et~al. 2023.
\newblock Llama: Open and efficient foundation language models.
\newblock \emph{arXiv preprint arXiv:2302.13971}.

\bibitem[{Vaswani et~al.(2017)Vaswani, Shazeer, Parmar, Uszkoreit, Jones,
  Gomez, Kaiser, and Polosukhin}]{vaswani2017attention}
Ashish Vaswani, Noam Shazeer, Niki Parmar, Jakob Uszkoreit, Llion Jones,
  Aidan~N Gomez, {\L}ukasz Kaiser, and Illia Polosukhin. 2017.
\newblock Attention is all you need.
\newblock \emph{Advances in neural information processing systems}, 30.

\bibitem[{Velickovic et~al.(2017)Velickovic, Cucurull, Casanova, Romero, Lio,
  Bengio et~al.}]{velickovic2017graph}
Petar Velickovic, Guillem Cucurull, Arantxa Casanova, Adriana Romero, Pietro
  Lio, Yoshua Bengio, et~al. 2017.
\newblock Graph attention networks.
\newblock \emph{stat}, 1050(20):10--48550.

\bibitem[{Wang et~al.(2022)Wang, Chen, Ding, Peng, Wang, Lin, Han, Hou, Li, Liu
  et~al.}]{wang2022maven}
Xiaozhi Wang, Yulin Chen, Ning Ding, Hao Peng, Zimu Wang, Yankai Lin, Xu~Han,
  Lei Hou, Juanzi Li, Zhiyuan Liu, et~al. 2022.
\newblock Maven-ere: A unified large-scale dataset for event coreference,
  temporal, causal, and subevent relation extraction.
\newblock \emph{arXiv preprint arXiv:2211.07342}.

\bibitem[{Wolf et~al.(2020)Wolf, Debut, Sanh, Chaumond, Delangue, Moi, Cistac,
  Rault, Louf, Funtowicz et~al.}]{wolf2020transformers}
Thomas Wolf, Lysandre Debut, Victor Sanh, Julien Chaumond, Clement Delangue,
  Anthony Moi, Pierric Cistac, Tim Rault, R{\'e}mi Louf, Morgan Funtowicz,
  et~al. 2020.
\newblock Transformers: State-of-the-art natural language processing.
\newblock In \emph{Proceedings of the 2020 conference on empirical methods in
  natural language processing: system demonstrations}, pages 38--45.

\bibitem[{Xiang and Wang(2023)}]{xiang2023survey}
Wei Xiang and Bang Wang. 2023.
\newblock A survey of implicit discourse relation recognition.
\newblock \emph{ACM Computing Surveys}, 55(12):1--34.

\bibitem[{Yuan et~al.(2023)Yuan, Huang, Cao, and Wen}]{yuan2023discriminative}
Changsen Yuan, Heyan Huang, Yixin Cao, and Yonggang Wen. 2023.
\newblock Discriminative reasoning with sparse event representation for
  document-level event-event relation extraction.
\newblock ACL.

\bibitem[{Zhao et~al.(2016)Zhao, Liu, Zhao, Chen, and Nie}]{zhao2016event}
Sendong Zhao, Ting Liu, Sicheng Zhao, Yiheng Chen, and Jian-Yun Nie. 2016.
\newblock Event causality extraction based on connectives analysis.
\newblock \emph{Neurocomputing}, 173:1943--1950.

\bibitem[{Zheng et~al.(2018)Zheng, Aragam, Ravikumar, and Xing}]{zheng2018dags}
Xun Zheng, Bryon Aragam, Pradeep~K Ravikumar, and Eric~P Xing. 2018.
\newblock Dags with no tears: Continuous optimization for structure learning.
\newblock \emph{Advances in neural information processing systems}, 31.

\bibitem[{Zuo et~al.(2021)Zuo, Cao, Chen, Liu, Zhao, Peng, and
  Chen}]{zuo2021improving}
Xinyu Zuo, Pengfei Cao, Yubo Chen, Kang Liu, Jun Zhao, Weihua Peng, and Yuguang
  Chen. 2021.
\newblock Improving event causality identification via self-supervised
  representation learning on external causal statement.
\newblock \emph{arXiv preprint arXiv:2106.01654}.

\bibitem[{Zuo et~al.(2020)Zuo, Chen, Liu, and Zhao}]{zuo2020knowdis}
Xinyu Zuo, Yubo Chen, Kang Liu, and Jun Zhao. 2020.
\newblock Knowdis: Knowledge enhanced data augmentation for event causality
  detection via distant supervision.
\newblock \emph{arXiv preprint arXiv:2010.10833}.

\end{thebibliography}

\clearpage

\appendix

\section{EventStoryLine Dataset Preprocessing}
\label{app:ESCPreprocessing}
Following the assumption that ``provided that pairs of events have a purely causal relationship, that is edges represent causal relations between the events, we will have a directed acyclic graph''~\citep{gopnik2007causal}. However, in the EventStoryLine dataset, not all document-level \textit{event causality graphs} (ECG) constructed with event nodes and ground truth causal relations as edges are directed acyclic graphs.
\subsection{Directed Acyclic Detection}
For a document, we can construct an ECG $\mathcal{G}$ with an \textit{adjacency matrix} $\mathbf{A}$: $\mathbf{A}{ij}=1$ if there exists a causal relation from $e_i$ to $e_j$; otherwise, $\mathbf{A}{ij}=0$, based on ground truth causal relations. We employ the directed acyclic constraint proposed by NOTEARS~\citep{zheng2018dags} on $\mathbf{A}$:
\begin{align}{\label{Eq:AcyclicConstraint}}
	\mathcal{H}(\mathbf{A})=tr(e^{\mathbf{A} \odot \mathbf{A}})-d=0
\end{align}
where $tr(\cdot)$ denotes the matrix trace operation, $\odot$ is the Hadamard product, and $d$ is the number of nodes. If the adjacency matrix $\mathbf{A}$ satisfies Equation~\eqref{Eq:AcyclicConstraint}, then $\mathcal{G}$ is a directed acyclic graph.
\par
Finally, we find that 16 documents in the EventStoryLine dataset are not satisfied with this constraint.
\subsection{Event Conflict Relation Detection}
We note that \citet{caselli2017event} only label 2,265 causal relations, and the rest are extended using within-document event co-reference chains. One possible reason is that the expansion process introduces false causal relations. Considering the mutually exclusive relation between co-reference and causality in event relations, we conduct a further analysis of the dataset. We observe that some co-referenced event pairs are incorrectly labeled as causality. Since the EventStoryLine dataset is composed by the Event Coreference Bank+ corpus, we remove causal relations that conflict with co-reference relations in all documents.

\begin{table}[t]
	\centering
	\resizebox{0.95\columnwidth}{!}{
			\renewcommand\arraystretch{1.2}
			\begin{tabular}{|l|c|c|}
				\hline
				\textbf{Item}	&	\textbf{Original Size}	&	\textbf{Preprocessed Size}\\
				\hline
				Topic	&	22	&	22\\
				\hline
				Documents	&	258	&	258\\
				\hline
				Sentences	&	4,316	&	4,316\\
				\hline
				Event Mentions	&	5,334	&	5,334\\
				\hline
				Intra-sentence causal links	&	1,770	&	1,751	\\
				\hline
				Cross-sentence causal links	&	3,855	&	3,727	\\
				\hline
				The Total causal links	&	5,625	&	5,478	\\
				\hline
		\end{tabular}}
		\caption{The EventStoryLine v0.9 dataset}
		\label{Tab:EventStoryLine}
	\end{table}

\subsection{Manual Check}
After Event Conflict Relation Detection, we note that there are still 6 documents that do not satisfy the directed acyclic constraint. We invite 3 ECI task researchers to check the causal relations in these 6 documents according to the annotated requirements~\citet{caselli2017event}, aiming to minimize causal loops in the ground truth causal relations. Table~\ref{Tab:EventStoryLine} presents the statistics of the EventStoryLine v0.9 dataset before and after preprocessing.

\section{Implementation Details}\label{Sec:DetailsAboutexperimentalSettings}
Our method is implemented based on the PyTorch version of Huggingface Transformer~\citep{wolf2020transformers}. We use the uncased BERT-base~\citep{devlin2018bert} as the base PLM and fine-tune it during the training process. We optimize our model using AdamW~\citep{loshchilov2018decoupled}, with a linear warm-up for the first 10$\%$ of steps. The learning rate for the PLM is set to 2e-5, while for other modules, it is set to 1e-4. The batch size is set to 1. The number of attention head $K$ is set to 4. The weights $\beta$ is set to 0.7. The relation confidence threshold $\omega$ is set to 0.6. We set the maximum iteration number $L$ to 9 for the EventStoryLine dataset and 19 for the MAVEN-ERE dataset. The structural difference threshold $\delta$ is set to 2. For loss function, we set the focusing parameter $\gamma$ to 2, the weighting factor of CAUSE/EFFECT class $\alpha$ to 0.75, and the weighting factor of NONE class is $1-\alpha$. The MLP is a two-layer fully connected network, in which the activation function is LeakyReLU and the rate of dropout is 0.4.
\par
The MAVEN-ERE primarily comprises lengthy documents with a large number of sentences. Capturing sufficient semantics using a single-sentence encoding approach is challenging. Therefore, for experiments on the MAVEN-ERE dataset, we employed the same encoding method as \citet{chen2022ergo}, leveraging dynamic window and event marker techniques. In addition, in the existence setting, to fulfill the requirement of constructing a directed event causality graph, we set all the positive samples to the CAUSE direction for the EventStoryLine dataset.




\section{Algorithm Complexity Comparison}
\label{app:AlgorithmComplexityComparison}
Table~\ref{Tab:AlgorithmComplexityComparison} compares the average resource consumption of various algorithms on the EventStoryLine (v0.9) dataset, considering both time and space perspectives on an NVIDIA RTX 3090 GPU with 24GB memory.
\par
From the Table, we observe that the directional causal graph generation and iterations for causality updates increase the time complexity of algorithm. However, due to the excellent identification performance of our model, we believe that the time cost is within an acceptable range, especially compared to the \citet{yuan2023discriminative}. Additionally, due to the adopted ECG updating approach, our method performs exceptionally well in terms of spatial complexity.
\par
We argue that the key to balancing algorithmic complexity and recognition performance lies in the actual number of iterations. Therefore, we propose termination criterion to achieve the trade-off between algorithm complexity and identification performance.
\begin{table}[t]
	\centering
	\resizebox{\columnwidth}{!}{%
			\renewcommand\arraystretch{1.6}
			\begin{tabular}{l|cc}
				\hline
				\textbf{Method}	&	\textbf{Time Per Epoch}	&	\textbf{GPU Per Batch}\\
				\hline
				\text{SENDIR} \citep{yuan2023discriminative}	&	2362 s	&	24 GB\\
				\text{ERGO} \citep{chen2022ergo}	&	\textbf{131 s}	&	24 GB\\
				\textbf{iLIF}	&	823 s	&	\textbf{16 GB}\\
				\hline
			\end{tabular}
		}
		\caption{Results of algorithm complexity comparison on the EventStoryLine dataset. Bold numbers represent the smallest cost.}
		\label{Tab:AlgorithmComplexityComparison}
	\end{table}
\section{LLaMA2 Prompt Details}
\label{app:LLaMA2PromptDetails}
We evaluate LLaMA2’s performance under zero-shot settings. Following previous works~\citep{gao2023chatgpt}, only the top 20 topics of EventStoryLine dataset are used for evaluation. For the event $e_i$ and event $e_j$, we design two prompt templates as follows:
\begin{itemize}
	\item \textit{causality existence prompt}
	\par Input: \{document content\}
	\par Question: is there a causal relationship between " \{$e_i$\} " and " \{$e_j$\} " ? Let's think step by step.
	\item \textit{causality direction prompt}
	\par Question: is the causal direction between " \{$e_i$\} " and " \{$e_j$\} " from " \{$e_i$\} " to " \{$e_j$\} "? Let's think step by step.
\end{itemize}
where we fill the templates with event mentions in the document to reference specific events.
\par
\begin{figure}[t]
	\centering
	\includegraphics[width=1.0\columnwidth]{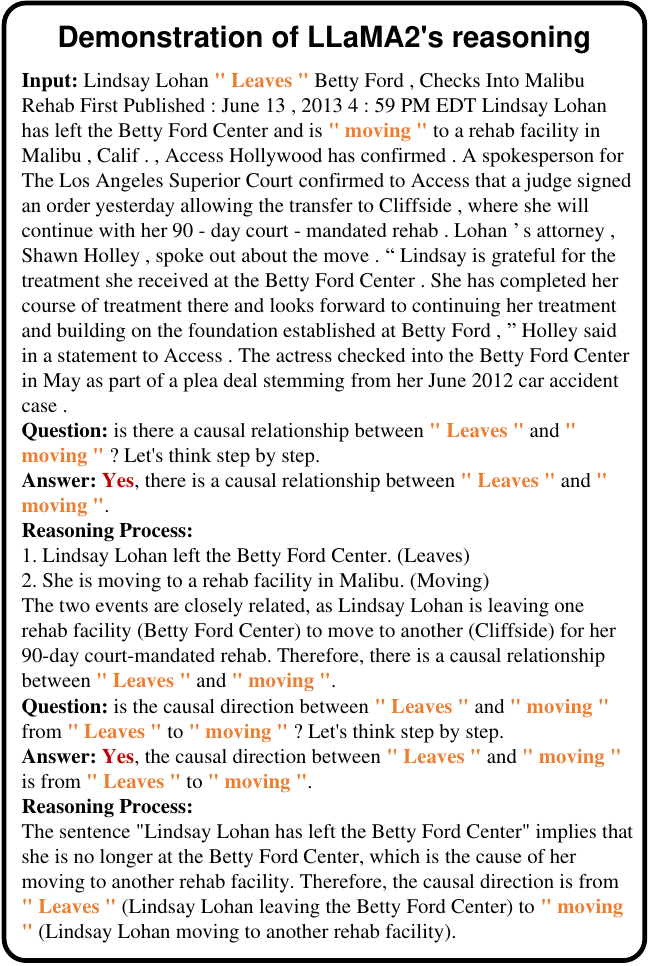}
	\caption{An example of LLaMA2 reasoning ECI task}
	\label{Fig:LLMs_prompt}
\end{figure}

 We first input the \textit{causality existence prompt} to obtain the \textit{existence identification} results. If there is a causal relation, we input the \textit{causality direction prompt} for continuous questioning to obtain the \textit{direction identification} results. Figure~\ref{Fig:LLMs_prompt} shows an example of LLaMA2 reasoning ECI task.
\end{document}